# Prioritized experience replay-based DDQN for Unmanned Vehicle Path Planning

Liu Lipeng [1],[*]
Peking University
Beijing, China
lipeng.liu@pku.edu.cn

Letian Xu[2]
Independent Researcher
lottehsu@gmail.com

Jiabei Liu[3]
Northeastern University
USA
liu.jiabe@northeastern.edu

Haopeng Zhao[4]
New York University
USA
hz2151@nyu.edu

Tongzhou Jiang[5]
Independent Researcher
tojiang0111@gmail.com

Tianyao Zheng[6]
Independent Researcher
tian971227@gmail.com

*Abstract*— Path planning module is a key module for autonomous vehicle navigation, which directly affects its operating efficiency and safety. In complex environments with many obstacles, traditional planning algorithms often cannot meet the needs of intelligence, which may lead to problems such as dead zones in unmanned vehicles. This paper proposes a path planning algorithm based on DDQN and combines it with the prioritized experience replay method to solve the problem that traditional path planning algorithms often fall into dead zones. A series of simulation experiment results prove that the path planning algorithm based on DDQN is significantly better than other methods in terms of speed and accuracy, especially the ability to break through dead zones in extreme environments. Research shows that the path planning algorithm based on DDQN performs well in terms of path quality and safety. These research results provide an important reference for the research on automatic navigation of autonomous vehicles.

Keywords- deep reinforcement learning; double DQN; path planning

## I. Introduction

In the field of unmanned vehicle planning, achieving efficient and accurate automatic decision-making has always been an important research direction. Especially in complex environments, unmanned vehicles must be able to respond quickly and make optimized decisions. In recent years, deep reinforcement learning (DRL) technology has shown great potential in unmanned vehicle path planning, especially the application of dual deep Q network (DDQN) in reducing overestimation errors.

Prioritized Experience Replay (PER), as an improved experience replay mechanism, significantly improves the efficiency and effect of learning by prioritizing important learning experiences for learning. Although the combination of DDQN plus PER has made progress in learning efficiency, it still faces challenges in processing high-dimensional data and real-time processing requirements. This prompts researchers to explore more efficient algorithms. For example, Reference [1] uses the enhanced dynamic window approach (DWA) algorithm, combined with dual-delay deep deterministic policy gradient (TD3), to significantly improve the robot's obstacle avoidance capability. This method significantly improves navigation efficiency and safety, and optimizes path planning and real-time obstacle avoidance functions.

In addition, hardware acceleration has become one of the key technologies to support real-time processing of complex robotic systems. Research in Reference [2] and Reference [3] has made significant progress in hardware-level real-time workload forensics, which provides an important reference for hardware optimization of robotic systems. At the same time, the research of Reference [4] greatly accelerated the training process of graph neural networks through extremely fast GPU kernel design, further supporting the efficiency of robots in processing large-scale graphics data.

In addition to advances in hardware and algorithms, data processing and augmentation technologies also play a key role in robot planning. The research of Reference [5] significantly improves the prediction ability in social network advertising through data augmentation technology. These technologies can also be applied to robot vision systems to improve the robot's ability to recognize and respond to the environment.

The research of Reference [6] proposed a bottom cruise trajectory planning algorithm for underwater robots. This research shows that accurate path planning can be achieved even in restricted spaces. Reference [7] explores a robust placement system based on detection and tracking for automatic aerial refueling of drones, showing the potential of deep learning technology in dynamic and uncertain environments.

Meanwhile, Reference [8] discusses the transition from assistive devices to swarms of collaborative robots in manufacturing and emphasizes autonomy and efficiency in group collaboration[9][10]. These studies reveal the possibility of solving complex tasks through advanced automation and intelligence[11][12].

Further, the research in Reference [13] achieved adaptive speed planning for unmanned vehicles through deep reinforcement learning. This research uses environmental feedback to dynamically adjust the decision-making process and optimize the robot's performance in uncertain environments [14]. Reference [15] further improves the accuracy of unsupervised high-resolution stereo matching in the field of visual systems through adaptive cost volume representation[16], which provides important technical support for robot visual perception and spatial positioning [17][18].

Combining the above research, this paper aims to explore a unmanned vehicle planning method suitable for complex environments by integrating DDQN and PER. This method not only needs to process real-time data and high-dimensional state space, but also ensures the accuracy and stability of decision-making while ensuring decision-making efficiency.

## II. METHODOLOGY

### A. Deep Reinforcement Learning Framework

Deep Q Network (DQN) is a reinforcement learning algorithm that combines Q learning and deep neural networks. Proposed by DeepMind, DQN has shown great potential in processing complex environments, especially in gaming environments (such as Atari) and achieved remarkable success. The main goal of DQN is to approximate the Q-value function by using deep neural networks, which can effectively handle high-dimensional state spaces.

In Q-learning, the goal is to learn a policy that maximizes the cumulative reward by approximating the optimal action-value function $Q^*(s,a)$. The update rules of Q learning are as follows (1):

$$Q(s_t,a_t) \leftarrow Q(s_t,a_t) + \alpha \left[ r_t + \gamma \max_{a'} Q(s_{t+1}, a') - Q(s_t, a_t) \right] \quad (1)$$

Where $s_t$ and $a_t$ are the state and action at time step $t$, $r_t$ is the reward received after taking action $a_t$ in state $s_t$, $\gamma$ is the discount factor, $\alpha$ is the learning rate.

Double Deep Q Network (DDQN) is an improvement on the traditional DQN algorithm and aims to solve the overestimation problem in Q learning. DDQN significantly reduces bias in Q-learning by decoupling action selection from Q-value evaluation.

Double Q-learning introduces two independent Q-value functions $Q_A$ and $Q_B$, with different parameters $\theta_A$ and $\theta_B$ respectively. The update rules are as follows (2):

$$Q_A(s_t,a_t) \leftarrow Q_A(s_t,a_t) + \\ \alpha \left[ r_t + \gamma Q_B(s_{t+1}, \arg\max_{a'} Q_A(s_{t+1},a');\theta_B) - Q_A(s_t,a_t) \right] \quad (2)$$

This double Q-learning method effectively reduces the bias caused by overestimation of a single Q-value function by using one Q-value function to select actions and another Q-value function to evaluate the action value.

In DDQN, action selection and Q-value evaluation are separated through two independent neural networks. The target value of DDQN is calculated as follows (3):

$$y_t^{DDQN} = r_t + \gamma Q(s_{t+1}, \arg\max_{a'} Q(s_{t+1}, a';\theta);\theta^-) \quad (3)$$

In this approach, the current Q-network $\theta$ is used to select the next action $a'$, while the target Q-network $\theta^-$ is used to evaluate the value of that action. The loss function of DDQN is (4):

$$L(\theta) = \mathbb{E}_{(s,a,r,s')}\left[ \left( y_t^{DDQN} - Q(s,a;\theta) \right)^2 \right] \quad (4)$$

This structure reduces the risk of overestimation and significantly improves the performance and stability of the algorithm by separating selection from evaluation.

We package the lidar data into 15 groups, and record the minimum value of each group as the distance from the obstacle and the position of the robot as the input of the neural network. The neural network consists of two fully connected (FC) layers of size 512*1. A rectified linear unit (ReLU) is activated after each layer. The output is a set of 25 actions consisting of five discrete speeds and turning angles.

### B. Prioritized experience replay

Prioritized Experience Replay (PER) is a technology in deep reinforcement learning that is used to improve the efficiency of the experience replay mechanism. Experience replay is a method of breaking data correlations by storing and randomly sampling past experiences, and PER further improves this method by assigning important experiences a higher sampling probability, thus improving more effectively. Learning efficiency and stability.

In reinforcement learning, an agent continuously collects experience through interaction with the environment. Typically, these experiences are stored in the form of four-tuples, i.e. A. The experience replay mechanism stores these quadruples and randomly samples small batches of experience during training to break the correlation between data and stabilize the training process. Traditional experience replay uses uniform sampling, but this can lead to underutilization of some key experiences.

PER assigns a priority to each stored experience and adjusts the sampling probability according to the priority. The higher the priority experience, the greater the probability of being sampled. This method can ensure that important experiences are used more times, thereby accelerating the learning process.

In PER, the priority of experience is usually calculated based on the Temporal-Difference (TD) error. The TD error reflects the difference between the current Q-value estimate and the actual return. The larger the error, the greater the contribution of the experience to updating the Q-value function, and the higher the priority is (5):

$$\delta_i = | r_i + \gamma \max_{a'} Q(s_{i+1}, a';\theta) - Q(s_i, a_i;\theta) | \quad (5)$$

$\delta_i$ is the TD error of the $i$ experience.

The priority $p$ of each experience is usually set to $\delta_i + \epsilon$, where $\epsilon$ is a small positive number to avoid a zero priority situation. The sampling probability $P(i)$ can be calculated by the following formula (6):

$$P(i) = \frac{p_i^\alpha}{\sum_k p_k^\alpha} \quad (6)$$

$\alpha$ controls the importance of priority. When $\alpha = 0$, it becomes uniform sampling.

Since PER changes the sampling distribution of experience, importance sampling weights need to be introduced to correct the sampling bias to ensure the unbiasedness of the estimate. The importance sampling weight $\omega$ is calculated as follows (7):

$$w_i = \left(\frac{1}{N} \cdot \frac{1}{P(i)}\right)^\beta \quad (7)$$

$N$ is the size of the experience buffer, $\beta$ controls the strength of importance sampling, usually increasing gradually from a small value to 1.

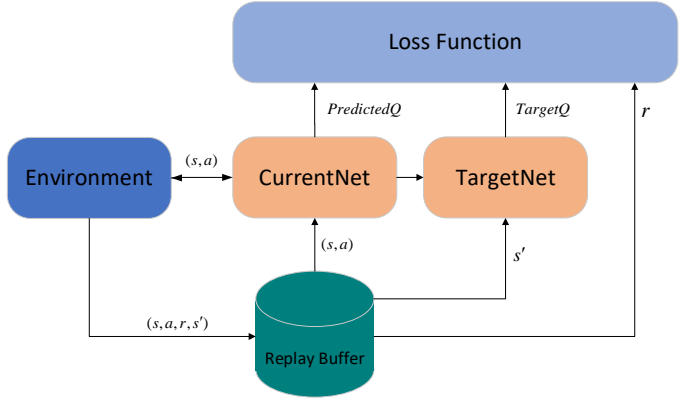

Fig 1. The whole process of DDQN

The overall framework of the algorithm is shown in Fig 1. We use $Q^*(s,a)$ in *CurrentNet* to select the optimal action, but when calculating $y_t^{DDQN}$, we use *TargetNet* to calculate and select the transition from the Replay Buffer to realize the training of unmanned vehicles.

We use the PER method to solve the problem of unmanned vehicles falling into dead zones and increase the weight of cases that cause planning failure in the training process.

## III. EXPERIMENTS

All experiments are conducted on a computer equipped with Xeon(R) E5-2690 2.6Ghz, 64Gb RAM and Nvidia GTX 1080ti, Python3.7, Torch 1.8.1, Cuda11.1. All experiments are based on the ubuntu20.04 operating system and ROS, and the physical simulation is performed in Gazebo.

During the training process, the Gazebo environment will issue instructions to the parameter server and assign random starting and target points to the unmanned vehicle. The unmanned vehicle's model will be redeployed and the data in the parameter server will be deleted to start a new round of training. During the training process, the lidar data collected in the Gazebo environment will be received in real time. This data will be used for training through the neural network and the output will be for the unmanned vehicle. control instructions.

Subsequently, the currently returned reward is evaluated based on the current state of the unmanned vehicle and relevant information provided in the Gazebo environment. If the unmanned vehicle reaches the target point, a large reward is provided and a new round of training begins.

The experimental parameters are shown in Table Ⅰ, which contains the main parameters involved in the experimental process. The simulation experiment environment is shown in Fig 2. We equipped the unmanned vehicle with lidar as sensing input.

TABLE I. TRAINING CONFIGURATION.

| parameter | Numerical value |
| --- | --- |
| $\varepsilon$-greedy $\varepsilon$ | 0.05 |
| Discount rate $\gamma$ | 0.99 |
| Learning rate $\alpha$ | 0.03 |
| Control degree $\beta$ | 0.1 |
| Number of iterations | 2500 |
| Replay buffer size | 3000 |

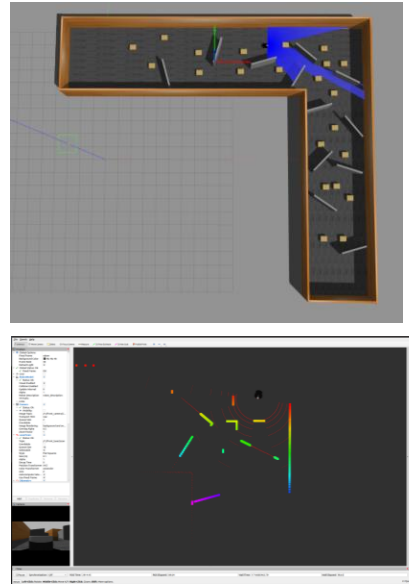

Fig 2. Planning environment

We compared four situations in the experiment, namely the original DQN algorithm, the DDQN algorithm, the original DQN algorithm combined with PER, and the DDQN algorithm combined with PER, and randomly selected 100 starting points and end points.

Figure 3 shows the growth process of cumulative rewards during the training process. It can be seen from the figure that the original DQN algorithm is difficult to get rid of the dead zone problem caused by the complex spatial environment and falls into local oscillations for a long time. Although the DDQN algorithm can alleviate the problem of excessive Q value, it does not perform well in the face of the dead zone problem. When combined with PER, the reward return of the DDQN algorithm is significantly higher than that of other algorithms.

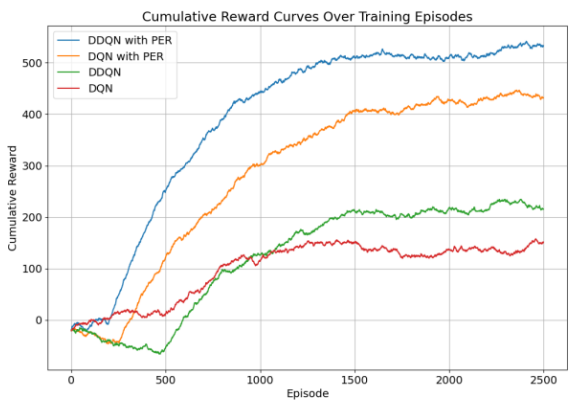

Fig 3. Cumulative reward during all the training episodes

Table Ⅱ shows the results of 100 random experiments. It can be seen from the table that the overall performance of the DDQN combined with PER algorithm in the planning process is better than all other algorithms and can successfully break through the limitations of environmental obstacles and various corners. case. The performance of the DQN algorithm combined with PER is slightly worse. Due to the overestimation of the Q value, the performance is not as good as DDQN when selecting actions. The DDQN and DQN algorithms perform relatively poorly in overall performance.

TABLE II. THE RESULTS OF 100 RANDOM EXPERIMENTS

| Method | Planning success rate(%) | Avg. Time (s) | Avg. Len (m) |
| --- | --- | --- | --- |
| DDQN with PER | 99 | 13.9 | 15.31 |
| DQN with PER | 97 | 13.4 | 15.87 |
| DDQN | 84 | 16.3 | 17.35 |
| DQN | 77 | 18.1 | 16.94 |

## IV. CONCLUSIONS

Unmanned vehicles often fall into dead zone problems when they enter extremely cramped spaces and are unable to escape the current environment for a long time, leading to final planning failure. This paper proposes a deep reinforcement learning method based on DDQN combined with PER, which is applied to the unmanned vehicle's path planning module. Experiments show that compared with the traditional DQN method, the method proposed in this article can solve the problem of unmanned vehicles frequently falling into dead zones and improve the success rate of overall planning. In addition, this method only requires the unmanned vehicle's lidar sensor as input, which greatly improves the universality of the method.